\documentclass[11pt]{article}
\usepackage{acl} 

\usepackage{times}
\usepackage{latexsym}
\usepackage{microtype}
\usepackage{graphicx}
\usepackage{amsmath,amssymb}
\usepackage{booktabs,multirow}
\usepackage{url}

\usepackage{placeins}
\usepackage{nccmath,amsmath,lipsum}
\usepackage{amssymb}
\usepackage{booktabs}
\usepackage{multirow} 

\usepackage[table]{xcolor}
\usepackage{graphicx}
\usepackage{array}
\usepackage{caption}
\usepackage{enumitem}

\title{Enhancing Job Matching: Occupation, Skill and Qualification Linking with the ESCO and EQF taxonomies}

\author{
\begin{tabular}{ccc}
Stylianos Saroglou & Konstantinos Diamantaras & Francesco Preta \\
\normalfont International Hellenic University    & \normalfont International Hellenic University    & \normalfont Tabiya \\
\normalfont \texttt{ssarglou@ihu.gr} & \normalfont \texttt{kdiamant@ihu.gr} & \normalfont \texttt{francesco.preta@tabiya.org}
\end{tabular}
\AND
\begin{tabular}{ccc}
Marina Delianidi & Apostolos Benisis & Christian Johannes Meyer \\
\normalfont International Hellenic University & \normalfont Tabiya & \normalfont University of Oxford \& Tabiya \\
\normalfont \texttt{dmarina@ihu.gr} & \normalfont \texttt{apostolos.benisis@tabiya.org}& \normalfont \texttt{christian.meyer@tabiya.org}
\end{tabular}
}

\begin{document}
\maketitle

\begin{abstract}
\noindent
This study investigates the potential of language models to improve the classification of labor market information by linking job vacancy texts to two major European frameworks: the European Skills, Competences, Qualifications and Occupations (ESCO) taxonomy and the European Qualifications Framework (EQF). We examine and compare two prominent methodologies from the literature: Sentence Linking and Entity Linking. In support of ongoing research, we release an open-source tool, incorporating these two methodologies, designed to facilitate further work on labor classification and employment discourse. To move beyond surface-level skill extraction, we introduce two annotated datasets specifically aimed at evaluating how occupations and qualifications are represented within job vacancy texts. Additionally, we examine different ways to utilize generative large language models for this task. Our findings contribute to advancing the state of the art in job entity extraction and offer computational infrastructure for examining work, skills, and labor market narratives in a digitally mediated economy. Our code is made publicly available: \url{https://github.com/tabiya-tech/tabiya-livelihoods-classifier}
\end{abstract}

\section{Introduction}

Recent developments in deep learning have spurred significant advancements in the job domain. This emerging field emphasizes the skill extraction paradigm, wherein deep neural networks are employed to extract skill-related information from plain-text job vacancies \cite{senger2024deep}. However, these texts also contain various other types of information—such as occupations and qualifications—that warrant further attention. Robust models should not only identify these additional entities but also link them to a knowledge base.
\newline
Linking job descriptions to established taxonomies—such as the European Skills, Competences, Qualifications and Occupations (ESCO) \cite{le2014esco} or the International Standard Classification of Occupations (ISCO)—remains a pivotal challenge. Early Large Language Models (LLMs) have demonstrated effectiveness in extracting robust semantic representations from unstructured text, as shown by \citet{devlin2018bert}. Building on this foundation, sentence embedding techniques introduced in SBERT \cite{reimers2019sentence} have further enhanced the efficiency of text classification and semantic similarity tasks, which are critical for mapping job descriptions to standardized occupational frameworks.

In this work, we address the following research question: How can language models be used to interpret and align the language of job advertisements with institutional classifications of labor and education, thereby offering insight into the structure and evolution of work?

We select the ESCO taxonomy as our use case and aim to match job vacancy texts to its nodes. This constitutes a text classification problem, for which we investigate two possible approaches.
\newline
\textbf{(Methodology 1)} We approach the task as a \textit{sentence linking} (SL) problem, feeding complete job descriptions into the models and expecting a list of ESCO nodes as outputs. This approach is often labeled as extreme multi-label classification \cite{decorte2023extreme, d2024context}.
\newline
\textbf{(Methodology 2)} We introduce an intermediate step where the models first perform entity recognition (ER) \cite{li2020survey}, thereby framing the task under the \textit{entity linking} (EL) paradigm \cite{sevgili2022neural}.

We explore both methodologies and present a comparative analysis using transformer-based neural networks as our foundation.

Previous studies have primarily concentrated on skill extraction from job vacancies, often overlooking other job-related entities. This limitation is likely due to the inherent complexity of the broader task. Defining what precisely constitutes a "skill" is itself challenging, introducing ambiguities into the training data.

Some prior work has adopted Methodology 1, applying plain sentence similarity strategies focused solely on skills. For example, \citet{khaouja2021unsupervised} compare using sent2vec trained on Wikipedia sentences with SBERT, which is trained on large collections of paraphrased sentences to generate embeddings. Similarly, \citet{zhang2022skill} employ language models to align n-grams extracted from job postings with the ESCO taxonomy. Furthermore, \citet{decorte2023extreme} and \citet{clavie2023large} utilize a synthetic skills training set to directly link sentences with skills, employing LLM-based re-rankers. In the work of \citealt{gnehm2022fine}, skill extraction is conducted directly by leveraging context-aware embeddings and the SBERT model, in a manner similar to \citet{zhang2022skill}. Moreover, their approach contextualizes skill domains within specific spans and ontology terms, utilizing ESCO's hierarchical structure. 

In contrast, Methodology 2 has not received as much academic attention.
An EL paper focused on the job domain was published by \citet{zhang2024entity}, in which the authors train two widely used models for this task: BLINK \cite{wu2019scalable} and GENRE \cite{de2020autoregressive}. They assess the effectiveness of skill extraction using this methodology with synthetic training data provided by \citet{decorte2023extreme}, achieving moderate yet promising results. The authors emphasize the need for a more comprehensive dataset for evaluation. In our work, we use the same evaluation set, introduced by \citet{decorte2022design}, for the skills component of our study.

SkillGPT \cite{li2023skillgpt} represents the first tool to employ a large language model (LLM) for the matching task. It transforms ESCO entries into structured documents, which the language model subsequently vectorizes. The input job text is then condensed into a summary, and the embedding of this summary is used to retrieve the most relevant ESCO entries. SkillGPT’s architecture resembles an EL pipeline, as it follows a two-step process. Although it incorporates both skill and occupation entities, the authors unfortunately do not provide an analytical evaluation.

Given the substantial progress in recent years, we aim to advance the research field by proposing an evaluation framework for skill, occupation, and qualification extraction concerning ESCO and EQF.

Data scarcity remains a significant challenge in the job domain when applying machine learning algorithms. To address this issue, we introduce three novel datasets: one for evaluating occupation linking with the ESCO taxonomy, another for qualification linking to the European Qualifications Framework (EQF), and a third for assessing occupation title similarity.
 

When considering a Retrieval-Augmented Generation (RAG) architecture \cite{gao2023retrieval}, both of the aforementioned methodologies can serve as the retrieval component of the system. Given the growing popularity of RAG and in-context learning \cite{d2024context, kavas2025multilingual}, it is essential to examine the respective strengths and weaknesses of these approaches. This constitutes the primary motivation behind our research.

We conducted extensive experiments on entity extraction using a well-established benchmark: the dataset introduced by \citet{green2022development}. We achieve state-of-the-art results on this benchmark.

For Occupation Extraction, SL proved to be the most effective approach, whereas EL yielded superior results for Skill Extraction. In our experimental setups, no single method emerged as clearly optimal for Qualification Extraction.



Finally, we explore a range of strategies for leveraging the LLMs to support our task; however, our findings do not indicate any substantial benefit from their use in this context.

List of common abbreviations for readability:
\begin{enumerate}[noitemsep]
\item \textbf{ESCO} - European Skills, Competences, Qualifications and Occupations
\item \textbf{EQF} - European Qualification Framework
\item \textbf{ER} - Entity Recognition
\item \textbf{SL} - Sentence Linking
\item \textbf{EL} - Entity Linking
\end{enumerate}
%

\section{Datasets}
\label{sec:datasets}
In this section, we present the datasets used throughout our work. These are categorized into three groups: reference sets, evaluation sets, and training sets.
\newline

%

\subsection{Reference Sets}
\noindent \textbf{ESCO}
The central aim of this study is to classify arbitrary English-language job vacancy texts using the ESCO taxonomy.

We utilize version 1.1.1 of ESCO, which contains 3,007 Occupations and 13,896 Skills. Both the Skill and Occupation frameworks are organized as taxonomies \cite{poli2010theory}—that is, they follow subclass relationships—where each Skill may have multiple parent categories. In this work, we focus exclusively on discrete entities within ESCO and disregard hierarchical relationships between broader concepts or links between Occupations and Skills. We leave this aspect for future exploration.



\noindent \textbf{EQF}
ESCO defines a qualification as the official outcome of an assessment by a competent body that verifies an individual’s learning achievements against established standards \cite{Q_ESCO}.
The qualification data available in Europass are sourced from national databases representing the frameworks of EQF member countries. Europass offers a consolidated repository of current, high-quality data on qualifications, national frameworks, and educational trajectories across Europe \cite{Q_Europass}.
We extract relevant information on EQF levels from the official European Union comparison portal.\footnote{\url{https://europass.europa.eu/en/compare-qualifications}} Only English-language content is retained. This results in a dataset of 814 entries, each consisting of a qualification string, the issuing country, and the corresponding EQF level (Table~\ref{tab:eqf_data_analysis}). 

\subsection{Evaluation Sets}
\noindent \textbf{Ethiopian Dataset}
To evaluate occupational classification, we employ a dataset comprising job descriptions annotated with corresponding ESCO occupation codes.

The vacancy data were collected from both online and offline sources in Ethiopia. Offline sources include physical job boards, public postings, and government gazettes across major cities. Online sources involve local job portals, an Ethiopian enterprise platform, and digital media managed by employers. Data are gathered either directly via the Ethiopian platform or through web scraping. In addition, printed job advertisements are photographed at the Ethiopian employment center for digital processing.

All collected data are reviewed and annotated by trained personnel using proprietary tools. Staff members receive specialized training on ESCO, ISCO, and O*NET classification systems, covering taxonomy structure, application rationale, and practical annotation exercises.

We compile real-world evaluation sets (Table~\ref{tab:evaluation_table}) for each entity type relevant to our models.

\noindent \textbf{Occupations}
We use a subset of the Ethiopian Jobs dataset containing 542 annotated entries (Table~\ref{tab:evaluation_table}), each comprising a job title, a job description, and the relevant ESCO occupation code. This subset is constructed ensuring diversity across multiple job sectors. The sensitive information in the original files has been redacted using Google DLP API.

\noindent \textbf{Skills}
For skill evaluation, we utilize the dataset introduced by \citet{decorte2022design}, which includes the HOUSE and TECH extensions of the SkillSpan dataset \cite{zhang2022skillspan}. These datasets feature test and development sets with SkillSpan entities mapped to the ESCO model.

\noindent \textbf{Qualifications}
We extend the Green Benchmark Qualifications dataset by mapping each entry to the appropriate EQF level.
 \begin{table}[!ht]
    \begin{center}
    \begin{tabular}{l|l|l|l}
    \toprule
    \textbf{Statistics}&\textbf{Occupations}&\textbf{Skills}&\textbf{EQF}\\
        \midrule
Data points& 542&920&448\\
 Avg entities& 1&2.7&1.3\\
 Avg words& 418.2&16.5&29\\
 Entities& 542&2406&595\\
 Words per entity& 3.4&3.1&3.4\\
 Max entities& 1&31&7\\
 Number of UNK& 0&981&361\\
        \bottomrule

    \end{tabular}
    \captionof{table}{\textbf{Evaluation Sets} Data Analysis with NLTK. Occupations and EQF evaluations sets are introduced in this study, while we use the aggregation of the TECH and HOUSE SkillSpan extensions provided by \citet{decorte2022design}.}
    \label{tab:evaluation_table}
\end{center}
\end{table}

\begin{table}[h]

    \centering
    \begin{tabular}{l|l}
        \toprule
          & \textbf{Statistics} \\
        \midrule
         Number of pairs & 210,175 \\
         Number of ESCO occupations & 1,156 \\
        \bottomrule
    \end{tabular}
        \captionof{table}{\textbf{Title Similarity Dataset} Data Analysis}
        \label{tab:hahu_trainset}
\end{table}


Two native Greek-speaking annotators (one male and one female) performed the annotation process. The resulting inter-annotator agreement, measured using Cohen’s Kappa \cite{fleiss1973equivalence}, was 0.45—indicating moderate agreement. Qualifications that did not align with any EQF level were labeled as unknown (UNK). A common example is the "driving license," which is not associated with any EQF level under ESCO.

To improve consistency, we resolve disagreements as follows: when both annotators select valid but differing EQF levels, we assign the lower level. If one annotator selects UNK while the other provides a valid EQF level, we consult \texttt{Gemini 1.5 Pro} as an adjudicator. The model is prompted to choose between the two annotations, and its decision is included in the final dataset. Details of the prompts used are provided in the Appendix. 

\subsection{Training Sets}
To support Methodology 2, we train entity extraction models using the benchmark dataset introduced by \citet{green2022development} (Table~\ref{tab:green_benchmark}).

\noindent \textbf{Title Similarity Dataset} To improve occupational classification performance, we further fine-tune two sentence transformers using a derived dataset based on the Ethiopian Dataset. This dataset (Table~\ref{tab:hahu_trainset}) is constructed by aligning job titles with the preferred and alternative labels defined in the ESCO occupation taxonomy. To ensure generalizability, test titles from the occupations evaluation set are excluded.
 





\section{Methodology \#1 : Sentence Linking}
\label{sec: LLM_apps}
\begin{table*}[!tb]
  \begin{center}
    \begin{tabular}{l|l|l|l|l|l}
        \toprule
\multicolumn{2}{l}{} &
\multicolumn{2}{c}{\textbf{Fulltext}} & \multicolumn{2}{c}{\textbf{Synthetic query}}\\
\midrule
& \textbf{Embeddings selection} & \textbf{mpnet}& \textbf{mpnet-ft} & \textbf{mpnet}& \textbf{mpnet-ft}\\
        \midrule
        \multirow{6}{*}{\rotatebox[origin=c]{90}{\textbf{\scriptsize Occupations}}} 
        & Single: Preferred Label &0.2416 & 0.4003 &0.2121  &0.3431\\
        
        & Single: Description & 0.3007&0.4889&0.2638 &0.3948\\

        & Single: Preferred Label and Description & 0.3339&\textbf{0.4981}&0.2841 &\textbf{0.4095}\\

        & Multiple: All fields & 0.2638&0.4298&0.2472 & 0.3892\\

        & Multiple: All fields separated  & 0.2675&0.3808& 0.2269& 0.3542\\
                
        \midrule
        \multirow{6}{*}{\rotatebox[origin=c]{90}{\textbf{\scriptsize Skills}}} 
        & Single: Preferred Label & \textbf{0.2211}&0.1785&\textbf{0.1820} &0.1544\\
        
        & Single: Description & 0.1208&0.0782& 0.0758&0.0689\\

        & Single: Preferred Label and Description & 0.1978&0.1607& 0.1268&0.1365\\
        
        & Multiple: All fields  & 0.2060&0.1909& 0.1544 & 0.1434\\

        & Multiple: All fields separated  & 0.2101&0.1744& 0.1475&0.1337\\
                
         \midrule
         \rotatebox[origin=c]{90}{\textbf{\scriptsize EQF}} & Single
         & \textbf{0.2881} & 0.2768& -&-\\
         \bottomrule
     \end{tabular}
\captionof{table}{\textbf{RAG-related Vector Search: sentence linking} The best results on each experiment are denoted in bold. Single indicates that only one embedding was generated for each target ESCO node, while multiple indicates than more than one embedding was generated. It is important to note that this evaluation is done on the sentence level.    The fulltext column refers to experiments done in the Methodology \#1 section and the synthetic query to the experiments of the Transformer Decoder Integration section.}
    \label{tab:LLM_vector_search}
\end{center}
\end{table*}

\begin{table*}[!tb]
\begin{center}
\begin{tabular}{l|l|l|l}
        \midrule
    \multirow{6}{*}{\rotatebox[origin=c]{90}{\textbf{\scriptsize Occupations}}} 
            & Single: Preferred Label & 0.3764&\textbf{0.5387}\\
            
            & Single: Description & 0.3321&0.4686\\
    
            & Single: Preferred Label and Description (concatenated) & 0.3339&0.4502\\
    
            & Multiple: All fields (separated) & 0.3782&\textbf{0.5387}\\
    
            & Multiple: All fields (separated secondary labels) & 0.3431&0.5000\\
                            
        \bottomrule
    \end{tabular}

    \caption{\textbf{RAG-related Vector Search: title linking} Experiment for occupation linking using job titles as queries in the Methodology \#1 section.}
    \label{tab:LLM_vector_search_titles}
    \end{center}
    
\end{table*}


Let $D$ be the Document space and a Sentence Transformer $ST : D \rightarrow \mathbb{R}_n$ be an embedding function to an arbitrary Euclidean metric space. Also, let $O  = \{ o_1, o_2, ..., o_{3007}\}$ , $S = \{ s_1, s_2, ..., s_{13896}\}$ and $Q = \{q_1, q_2, ..., q_{814}\}$ be the reference sets described in the previous section. Our goal is to retrieve entities from these sets so we embed $O, S$ and $Q$ using $ST$ and cache them in separate vector databases. We define a \textit{query} which is a plain-text sentence, annotated with entities from the reference sets.

We consider different possible ways of embedding the ESCO occupations and skills nodes and of comparing the embedding to the query, to find the one that maximizes precision. With respect to the Skills and Qualifications Evaluation Sets, we remove the UNK labels and link each sentence only one time. 

To improve Occupational matching, provided the title similarity dataset, we fine-tune the \texttt{all-mpnet-base-v2} sentence transformer. The model was trained on
minimizing the Multiple Negatives Loss \cite{henderson2017efficient}, using hyperparameters, as suggested by the sbert documentation \footnote{https://sbert.net/}.



The relevant textual fields for each ESCO node are: \textit{preferred label}, \textit{description}, \textit{secondary labels}, i.e., alternative titles presented as a newline-separated list.

To evaluate various embedding strategies, we consider the following configurations:
(1) Single embedding: \textit{preferred label}
(2) Single embedding: \textit{description}
(3) Single embedding: concatenation of \textit{preferred label} and \textit{description}
(4) Multiple embeddings: one per field (\textit{preferred label}, \textit{description}, and combined \textit{secondary labels})
(5) Multiple embeddings: one for \textit{preferred label}, one for \textit{description}, and individual embeddings for each \textit{secondary label}.

For retrieval, we calculate the cosine similarity between the embedded $query$ and each sentence, returning the top-$k$ most similar results. Evaluation is based on Accuracy@1, in line with prior work \cite{zhang2024entity, zaporojets2022tempel}. This metric is selected because the majority of examples in the evaluation set contain a single relevant item. In the case of Skills and Qualifications, however, some sentences may correspond to multiple relevant items. Therefore, the evaluation focuses on the models' ability to correctly identify at least one relevant item per instance.

For occupations, we find that the Single embedding: concatenation of preferred label and description strategy yields the best overall performance (0.4981), as it integrates information from both the title and the job description, thereby providing richer contextual representation. For Skills, using only the preferred labels is found to be the most effective approach (0.2211), as including a larger set of alternatives introduces noise into the database and negatively impacts performance. Finally, for Qualifications, we observe that this unsupervised approach does not produce optimal results (0.2881).


Fine-tuning on occupation-specific data significantly improves accuracy for the Occupations task, without negatively impacting performance on Qualifications. However, for Skills, we observe a drop in performance after fine-tuning, suggestive of catastrophic forgetting.

Next, we investigate whether an ER approach outperforms full-sentence embeddings. As a preliminary analysis, we rerun experiments on the Occupation evaluation sets using job titles (Title Linking) as queries, and compare their embedding-based retrieval performance against the earlier configurations. We observe a substantial improvement in performance, with an approximate 4\% increase (0.5387) in accuracy. The relatively low performance on Skills and Qualifications, combined with the improvements observed in Title Linking, strengthens our motivation to develop a dedicated EL model aimed at surpassing the current SL baseline.

Detailed results discussed in this section are presented on Tables \ref{tab:LLM_vector_search} and \ref{tab:LLM_vector_search_titles}.

\section{Methodology \#2: Entity Linking}
\label{sec:linking}

Given an input text document
$P = \{w_{1},..., w_{r}\}$ 
and a list of entity mentions (n-grams corresponding to entities)
$M_{P} = \{m_{1}, . . . , m_{n}\}$, the output of an EL model is a list of mention-entity pairs 
$\{(m_{i}, e_{i})\}, i\in[1,n]$. 
Each entity $e_i$ is an element in a set $E$ of all possible entities in a knowledge base (e.g. WikiData, DBpedia, ESCO).

Most EL-related works hypothesize that the mentions are explicitly given in the training and test datasets.
Inspired by \citet{sevgili2022neural}, we distinguish the mention detection and entity disambiguation steps and assume that the mention boundaries are missing from the evaluation procedure. 
Consequently, this model consists of two discrete modules.




\noindent \textbf{Entity Recognition Module} Formally, the ER task according to \citet{zhang2022skill} is defined as follows. Let $d$ be a subset of sentences (sequences of tokens) from a job posting $P$. Let $X^i_d =\{x_1, x_2, ..., x_T \}$ be the $i^{th}$ sequence of input tokens  and $Y^i_d = \{y_1, y_2, ..., y_T \}$ be the target sequence of BIO labels (e.g., “B-Skill”, “I-Occupation”, “O”) corresponding to this input sequence. The goal is to use $P$ to
train a sequence labeling algorithm $h : X \rightarrow Y $ to accurately predict entity spans by assigning an output label $y_t$ to each token $x_t$.

We perform the ER task by training BERT-based models for token classification. We experimented with language models of various sizes and pre-training schemes. Namely, we used \texttt{BERT} with both its base and large variances on the cased version. Also, we experimented with two domain-adapted models, \texttt{JobBERT} \cite{zhang2022skillspan} and \texttt{ESCOXLM-R} \cite{zhang2023escoxlm} to test whether domain adaptation generalizes in our holistic overview of job postings text analysis. Both \texttt{RoBERTa\textsubscript{base}} and \texttt{RoBERTa\textsubscript{large}} \cite{liu2019roberta} were fine-tuned on our task, as well as the first version of Microsoft's \texttt{DeBERTa\textsubscript{base}}\cite{he2020deberta} model.

Based on previous work \cite{zhang2022skillspan, souza2019portuguese, jensen2021identification}, we experimented adding a conditional random field \cite{lafferty2001conditional} decoder on top of transformer language models for improved accuracy.

\noindent 
\textbf{Entity Similarity Module}
Let $(m_i, e_i)$ be a tuple of an extracted mention by the entity extractor, where $m_i\in \mathcal{P}(D)$, $e_i \in E$. $D$ is the Document space, $\mathcal{P}(D)$ it's power set and $E$ is the set of entity categories.
Similar to SL, we represent Occupations $O$, Skills $S$, and Qualifications $Q$ using a Sentence Transformer (ST) to generate the corresponding embedding vectors in $\mathbb{R}^n$ space. Note that $E = O\cup S \cup Q$.


Given a job posting $P = \{w_1, ..., w_n\}$, we apply the NLTK \cite{bird2009nltk} package to tokenize the document into chunks, $X_k = \{ x_1, ..., x_k\}$. Each $X_k$ is passed through the ER function $h$ to generate the BIO labels $h(X_k) = Y_k = \{ y_1, ... y_k\}$. From these we can obtain the mentions $m_i$ and apply post-processing steps to improve performance.

These steps include: (1) removing special tokens (e.g. [SEP], [CLS], $<s>$, etc), (2) correcting common sequence errors such as converting the sequence (..., "B-", "O", "I-",...) to (..., "B-", "I-", "I-",...), and (3) ignoring single "I-" tags appearing at the end of a sentence. 

For each mention $m_i$ the Sentence Transformer is used to generate the embedding vector $\mathbf{V} = ST(m_i)\in \mathbb{R}^n$.

We then proceed to compute the cosine similarity of $\mathbf{V}$ against $o_j \in O$, $s_i \in S$ and $q_k \in Q$, depending on the category indicated by the ER module.
Finally, we retrieve ranked lists of the top-$k$ ESCO Occupations, ESCO Skills or EQF Qualification entities based on the above metric. 

We experiment with two sentence transformers: \texttt{all-MiniLM-L6-v2} and \texttt{all-mpnet-base-v2}. The \texttt{all-MiniLM-L6-v2} model is further fine-tuned on the title similarity set, similar to \texttt{all-mpnet-base-v2}. All four resulting models are evaluated as candidates for our sentence transformer function $ST$.

\subsection{Evaluation}
\label{sec:eval}

The ER training was assessed using standard span F1 strict metric \cite{li2020survey, nakayama2018seqeval}, where true positives are considered if the exact entity span is predicted. 
 
The entity similarity evaluation can be categorized into \textit{in-KB Evaluation} when all the entities in the evaluation set are from the same knowledge base, and \textit{out-of-KB Evaluation} when Unknown labels correspond to entities in the text. For the purposes of this study, we opt to report only the in-KB case for a more comprehensive evaluation. So, as we did with SL, we remove the UNK labels from the reference sets and perform the EL procedure. 

We developed an algorithm using the whole system to evaluate the similarity module. Specifically, based on the extracted entities on a given evaluation set, we check whether an overlap exists with the ground truth entity using the Jaccard Similarity \cite{jaccard1912distribution}. The ground truth span that maximizes the Jaccard Similarity with the extracted entity is then attributed to the top-$k$ retrieved entities from the reference sets. If no overlap exists, the system returns the UNK label. We evaluate the Entity Similarity Module based on Accuracy@1.




\subsection{Experiments}
\begin{figure}[h]
    \centering
    \includegraphics[width=1\linewidth]{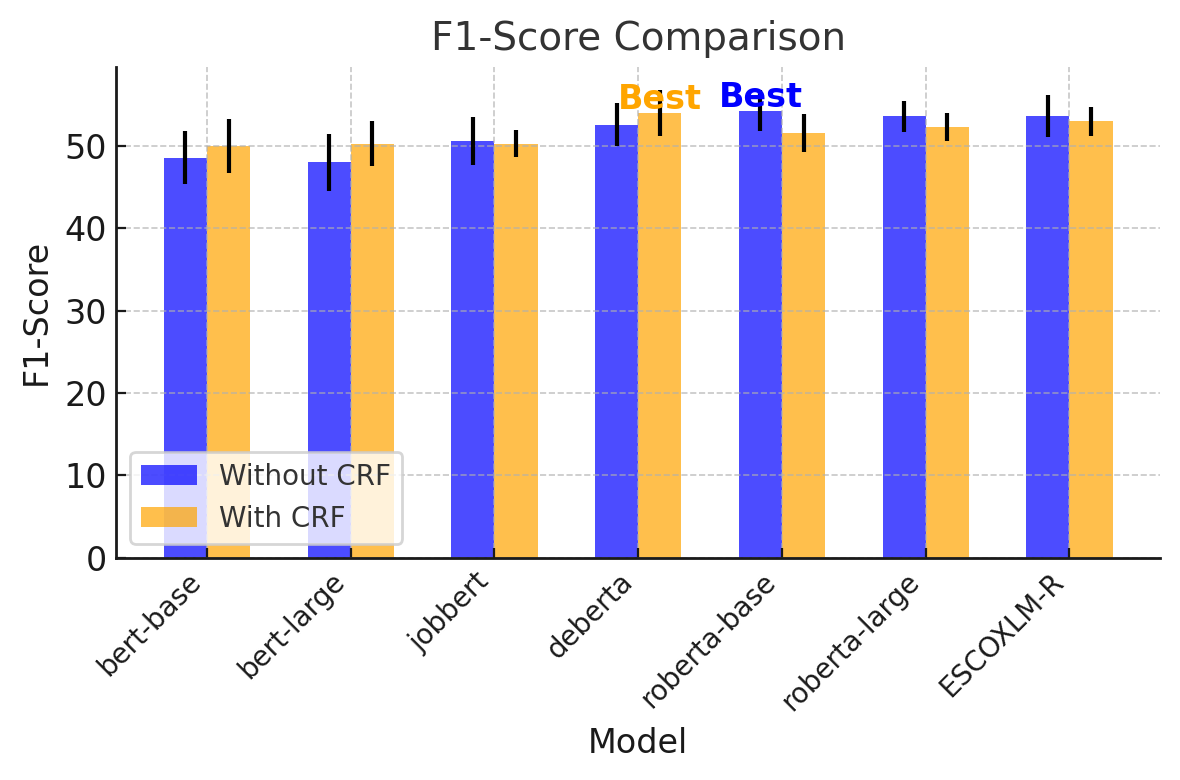} 
    \caption{F1-Score results on the Green Benchmark test set. The results show the mean and standard deviation over three random seeds.}
    \label{tab:entity_recognition_results}
\end{figure}

\noindent \textbf{Entity Recognition} This system's foundation is the ER module, which acts as the mention detector in the EL framework. Similar to traditional EL models \cite{sevgili2022neural,stern2012joint}, the ER errors propagate to entity disambiguation. 


All training was conducted using V100 GPUs provided by the Advanced Research Computing (ARC) of the University of Oxford. For the ER training, we performed a comprehensive grid search over hyperparameters for all encoder models. In all experiments, we employed the test set from \citet{green2022development} for both validation and evaluation, as a validation one was not introduced in the original study. The test set was the most carefully curated among the available sets \cite{green2022development}, ensuring both consistency and diversity. The best-performing configurations were selected as outlined in the Appendix.

Figure~\ref{tab:entity_recognition_results} presents the results of the entity extraction experiments conducted on the Green Benchmark dataset. Each model was trained using three different random seeds to ensure robustness. The optimal model for this dataset is \texttt{RoBERTa\textsubscript{base}}, which achieves a strict F1-score of \textbf{54.3}$\pm$2.6.

Overall, we observe that the addition of a CRF decoder enhances the performance of both \texttt{BERT} and \texttt{DeBERTa} models, but does not yield improvements for \texttt{RoBERTa}. 

Our best-performing model, \texttt{RoBERTa\textsubscript{base}}, establishes a new state-of-the-art on this benchmark. In our experimental setup, the previously reported state-of-the-art model, \texttt{ESCOXLM-R} \cite{zhang2023escoxlm}, achieves an F1-score of $53.6 \pm 2.5$ F1-score. 
\begin{table*}[t]
\begin{center}

    \begin{tabular}{l|l|l|l}

        \toprule
\textbf{Entity Type}& \textbf{Entity Model} & \textbf{Similarity Model}& \textbf{Accuracy@1}  \\
       \midrule
        Occupations     
        & roberta-large+CRF & all-mpnet-base-v2-FT & 0.489 \\
        
        \midrule
        Skills 
        & roberta-base+CRF & all-MiniLM-L6-v2 & 0.326\\
        \midrule
        EQF
        & bert-large-cased  & all-MiniLM-L6-v2 &0.350\\
\bottomrule
    \end{tabular}
\end{center}
    \captionof{table}{\textbf{Entity Linking Results} With FT (fine-tuned), we note the models that were fine-tuned on the Title Similarity Set. The evaluation for EL is performed on the entity level, where we treat as queries the extracted entities of the ER module.}
    \label{tab:linking_results}

\end{table*}

\noindent \textbf{Entity Similarity} We represent ESCO nodes using only preferred labels, as the extracted entities serve as the primary queries of interest. Introducing a larger number of reference items has been shown to introduce noise into the data, as observed in the SL analysis. The evaluation sets are assessed using fourteen fine-tuned entity extractors in combination with four sentence transformers. We report the best-performing models for each entity type in Table \ref{tab:linking_results}.


The EL procedure functions as a more comprehensive classification approach by enabling the extraction of richer information from the Knowledge Base, enabling more advanced engineering. However, the effectiveness of this methodology is highly dependent on the performance of the Entity Extraction module. When the model fails to identify the relevant entities, its ability to retrieve corresponding items is significantly impaired. Although this characteristic may offer potential advantages for out-of-KB evaluation, it may limit the overall capabilities of the model. 
At first glance, a substantial improvement in performance can be observed for Skills and Qualifications, accompanied by a slight decline for Occupations. However, due to the differing levels of evaluation, caution must be exercised when interpreting these results and drawing conclusions.



\section{Methodologies comparison} 
\label{sec:comparison}
 We must denote that the evaluation on SL is done on the sentence level, while the EL is done on the entity level, so it is not in one-to-one correspondence \cite{zhang2024entity}. For example, in the case of the Skills evaluation set, as shown in Table~\ref{tab:evaluation_table}, there are 920 possible queries at the sentence level, compared to 2406 at the entity level 920. Again, we do not include UNK classification labels in this comparison, removing them from the evaluation sets.  
 
To enable a fair comparison between these methods, the outputs of the EL system must be adjusted by aggregating the recommended entities into a single list, similar to the sentence linking approach. This transformation allows the entity-level evaluation to be mapped to the sentence level, thereby facilitating a direct comparison between the proposed methods. 

All EL experiments utilize the \texttt{RoBERTa\textsubscript{base}} model as the Entity Extractor. For Occupations, we employ the fine-tuned \texttt{all-mpnet-base-v2} sentence transformer, whereas the non-fine-tuned version is used for Skills and Qualifications. In all experiments, Preferred Labels are used as retrieval items, except in the Sentence Linking approach for Occupations, where the concatenation of the preferred label and description is used.  
\begin{table}[h]
\begin{center}
\begin{tabular}{l|l|l}
 \toprule
& \textbf{Method} & \textbf{Accuracy@1}\\
 \midrule

             \multirow{3}{*}{\rotatebox[origin=c]{90}{\textbf{\scriptsize Occupations}}}
             & Entity Linking & 0.4704\\ [0.08cm]
        
        & Sentence Linking  & 0.4981\\ [0.08cm]
        & Title Linking &\textbf{0.5387}\\ [0.08cm]
        
        \midrule
         \multirow{2}{*}{\rotatebox[origin=c]{90}{\textbf{\scriptsize Skills}}}
        & Entity Linking & \textbf{0.3969}\\
        & Sentence Linking & 0.2211\\   
         \midrule
         \multirow{2}{*}{\rotatebox[origin=c]{90}{\textbf{\scriptsize EQF}}}
        & Entity Linking & \textbf{0.2881}\\
        & Sentence Linking & \textbf{0.2881}\\      
       \bottomrule

\end{tabular}
    \captionof{table}{\textbf{Retrieval comparison}: The best-performing experiments for the Occupation, Skill, and Qualification reference sets—evaluated at the sentence level—are highlighted in bold.}
\label{tab:rag_comparison}
\end{center}
\end{table}

In Table \ref{tab:rag_comparison} we present a brief summary of the comparable results discussed so far, while in Table \ref{tab:qualitative_examples} we present some examples indicating the performance of our different approaches.

For \textbf{Occupation Extraction}, SL continues to outperform EL, primarily due to its ability to contextualize relevant information from the job description more effectively. Nevertheless, linking based solely on job titles yields the best results and is considered the most effective strategy when the job title is available within the job description.

For \textbf{Skill Extraction}, EL demonstrates a clear advantage over SL. Incorporating the broader context of the entity introduces noise into the process, whereas linking based exclusively on the relevant entity text yields the best results. 

Finally, for \textbf{Qualification Extraction}, similar results are observed in our evaluation set. The qualitative analysis does not reveal a clear advantage for either method, which may be attributed to the specific terminology used for qualifications in the UK job market. For future research, it is recommended to expand the evaluation set to include qualifications from more diverse and general job markets. As illustrated in Table~\ref{tab:qualitative_examples}, EL tends to underperform in cases where there is no lexical overlap between the extracted entity and the reference label.

\section{Transformer Decoder Integration}
\label{sec:decoders}

\begin{table*}[ht]
\centering
\begin{tabular}{p{0.5cm}|p{5.5cm}|p{4cm}|p{4cm}}
\toprule
\textbf{} & \textbf{Sentence} & \textbf{Sentence Linking} & \textbf{Entity Linking} \\
\midrule
1 & \textcolor{blue}{\textbf{Senior Trade Service Officer}} Job Requirement Education (Minimum): BA in Accounting... & \textcolor{green}{\textbf{trade development officer (2422.18)}} & \textcolor{red}{\textbf{procurement support officer (3343.1.5)}} \\
\midrule
2 & \textcolor{blue}{\textbf{Subject Specialist}} Teach one or more subjects to students at the secondary school level. Job Requirement Qualification: ... & \textcolor{green}{\textbf{secondary school teacher (2330.1)}} & \textcolor{red}{\textbf{securities analyst (2413.1.4)}} \\
\midrule
3 & A vacancy exists for a $<$PROFESSION$>$ $<$PROFESSION$>$ with strong \textcolor{blue}{\textbf{Java skills}} and experience of \textcolor{blue}{\textbf{working in the web layer}} working as a key member of an agile team. & 1. \textcolor{red}{\textbf{provide assistance with job search}} \newline
2. \textcolor{red}{\textbf{recruit employees}}& 1. \textcolor{green}{\textbf{Java (computer programming)}} \newline
2. \textcolor{green}{\textbf{web programming}}\\
\midrule
4 & As a person you are \textcolor{blue}{\textbf{innovative}} and \textcolor{blue}{\textbf{solution-oriented}}. & 1. \textcolor{red}{\textbf{show entrepreneurial spirit}} \newline
2. \textcolor{green}{\textbf{think innovately}}& 
1.\textcolor{green}{\textbf{think innovately}} \newline2.\textcolor{green}{\textbf{create solutions to problems}}
\\
\midrule
5 & An \textcolor{blue}{\textbf{HND / professional qualification}} at Level 4 and above in a related discipline Evidence of ongoing professional updating and development. &\textcolor{green}{\textbf{EQF4}} & \textcolor{red}{\textbf{EQF3}} \\ 
\midrule
6 & Able to work both individually and as part of a team ... Qualifications Good \textbf{\textcolor{blue}{university degree}}, in electronic engineering... &\textcolor{red}{\textbf{EQF5}} & \textcolor{green}{\textbf{EQF6}}  \\
\bottomrule
\end{tabular}
\caption{\textbf{Qualitative Analysis on SL vs. EL}. Annotated entities are marked in \textcolor{blue}{\textbf{blue}}, correct model predictions are shown in \textcolor{green}{\textbf{green}}, and incorrect predictions are indicated in \textcolor{red}{\textbf{red}}. Sentences 1 and 2 are drawn from the Occupations evaluation set, sentences 3 and 4 from the Skills evaluation set, and sentences 5 and 6 from the Qualifications evaluation set.  } 
\label{tab:qualitative_examples}
\end{table*}

In this section we explore different avenues of integrating the latest generation of Large Language Models (LLMs) for the task of linking sentences to the ESCO taxonomy.

From our experimentation, we concluded that linking job descriptions to ESCO with LLMs, like GPT-4 \cite{achiam2023gpt} and Gemini \cite{team2023gemini} directly was impossible at the time. It requires an understanding ESCO's hierarchical structures and precise concept definitions \cite{le2014esco}, where LLMs often produce hallucinations regarding the exact ESCO codes/ labels.

For a thorough evaluation, we opt to perform the ER task using a general-purpose decoder, the  \texttt{Gemini 1.5 Pro} model and an open-source one, the \texttt{Universal-NER} \cite{zhou2023universalner} model where the authors fine-tuned Llama \cite{touvron2023llama} to task-adapt it for ER and to output JSON format strings. We use the same prompt template for both models, the one suggested by \citet{zhou2023universalner}. We measure the performance of the models in terms of strict F1-score in the \citet{green2022development} benchmark. \texttt{Gemini 1.5 Pro} achieves 0.22 with one-shot prompting and 0.25 with five-shots. \texttt{Universal-NER} reaches 0.33. Both models, severely underperform supervised methods. 

Previous studies \cite{nguyen2024rethinking, wang2023gpt} have consistently shown that supervised approaches substantially outperform decoder-only models in terms of classification accuracy and consistency. These findings underscore the importance of domain-specific, fine-tuned decoders \cite{herandi2024skill} over reliance on in-context learning alone \cite{nguyen2024rethinking}. Nonetheless, transformer-based decoders have demonstrated utility \cite{decorte2023extreme, clavie2023large} in re-ranking the outputs of retrieval models—an avenue not explored in the present work.

On the other hand, one of the most prominent uses of transformer decoders is their ability to create synthetic data \cite{clavie2023large}. Inspired by the work of \cite{li2023skillgpt}, where they summarize the job description before performing similarity, using \texttt{Gemini 1.5 Pro}, we generate a new query from each sentence in Occupation and Skill evaluation sets. We prompt the model to produce sentences comparable to what a user with the given skill or occupation would tell the model when asked to describe their skills or occupation. Then, we embed such queries using \texttt{all-mpnet-base-v2} and its fine-tuned version. Detailed experiments can be found in table \ref{tab:LLM_vector_search}. We do not observe any improvements in the results. Possibly, there exists a prompt transformation that enhances information retrieval. This is a promising direction of future research \cite{d2024context, jiang2025deepretrieval}

\section{Conclusion}

In this work, we investigated optimal strategies for leveraging language models to link job vacancy texts to the ESCO taxonomy. Emphasizing the use of open-source models, we compared two main approaches: sentence linking (SL) and entity linking (EL), with the latter incorporating an entity recognition (ER) component. 

We introduced two novel datasets to support the evaluation of occupation and qualification extraction tasks, thereby broadening the focus beyond skill extraction for ESCO. These datasets are human-annotated with their respective ESCO entity type and comply with real-world job descriptions. Also, we release a dataset to support Occupation Title Similarity tasks. We acknowledge that the relatively low annotation agreement in the Qualification evaluation set may limit the dataset’s usefulness. As shown in Example 6 in Table \ref{tab:qualitative_examples}, determining the correct EQF level is not always straightforward. With this publication, we aim to raise awareness of these annotation challenges and encourage further research on how to overcome them.

Our findings underscore the complexity of mapping job-related entities to ESCO and EQF classifications. We conclude that applying SL with embedding strategies that incorporate rich contextual information yields the best results for Occupation Extraction. Conversely, for Skill Extraction, where contextual information provides limited benefit, the EL approach proves markedly more effective. In the case of Qualification Extraction, no method demonstrates a clear advantage, suggesting that a supervised classification approach may be most appropriate, contingent on the availability of suitable training data. Future systems may benefit from hybrid or constrained retrieval, built upon these two methods.



For ER we achieved state-of-the-art performance on the Green Benchmark Dataset \cite{green2022development} for entity extraction, attaining an F1 score of 54.3—surpassing the previous best of 51.2 reported by \citet{zhang2023escoxlm}.

Counterintuitively, our findings indicate that generative LLMs do not offer immediate improvements, although this direction warrants further exploration.

Through this work, we aim to lower the entry barrier for researchers and practitioners in the job domain interested in the job description classification task. To support continued progress in this area, we publicly release our codebase as a resource for the community.

\section*{Limitations}

\textbf{Data Diversity and Language}
This research was done primarily on English-speaking datasets, which could limit its effectiveness in job markets with diverse linguistic profiles.  Expanding handle multiple languages is recommended for future research.
Additionally, the ESCO framework is designed for Europe and may not capture precisely the low- and middle-income countries' job market. 
Perhaps other (a few) Occupations exist in their countries that do not exist in the ESCO.
Each specific country context has limitations, such as idioms used to refer to occupations or specially named qualifications. There exists ongoing research regarding this topic \footnote{https://docs.tabiya.org/overview}. 


\noindent \textbf{No Joint Training}
The lack of a comprehensive, AIDA-style \cite{hoffart2011robust} dataset tailored for entity linking job descriptions to taxonomies like ESCO presents a significant limitation. Existing datasets fail to capture the variability and context-dependent nature of job-related terminology and they focus on different kinds of entites. This deficiency hinders the development and evaluation of robust entity linking models, particularly those designed for joint training across diverse job domains.


\noindent \textbf{Closed-Source Models}
With the exception of Gemini, all models used in this study are open-source. While closed-source models have demonstrated superior performance in various scientific studies, we intentionally prioritized open-source alternatives to ensure transparency, reproducibility, and accessibility. This choice may have resulted in a trade-off in terms of maximum achievable performance.


\section*{Acknowledgement} The authors would like to acknowledge the use of the University of Oxford Advanced Research Computing (ARC) facility in carrying out this work (http://dx.doi.org/10.5281/zenodo.22558) and the hahu.jobs Ethiopian website for the provision of training and test data.

\bibliography{custom}

\clearpage
\appendix

\section*{Appendix}

\subsection*{Training Hyperparameters}
\label{sec:hyper}
In Table \ref{tab:model_size}, we present the model parameters that were used during this research.  
\begin{table}[h]
\centering

\begin{tabular}{l|c}
\hline
\textbf{Model Sizes} & \textbf{Parameters} \\
\hline
roberta-base & 124M \\

roberta-large & 354M \\

bert-base-cased & 107M \\

bert-large-cased & 332M \\

deberta & 138M \\

\hline
all-mpnet-base-v2 & 104M \\
all-MiniLM-L6-v2 &  22M \\
\hline
\end{tabular}
\caption{Model Sizes and Parameters}
\label{tab:model_size}
\end{table}

\subsubsection*{Entity Recognition}

For the ER training we did a hyperparameter search regarding the parameters batch size, epochs, and learning rate. We set the max length of the sentences to 128 tokens weight decay to 0.01, while we search from possible options batch size:{16, 32, 64}, epochs:{5, 10}, learning rate: {0.0001, 0.00005, 0.00001}. In Table \ref{tab:erhyper}, we present the best hyperparameters with respect to the results in Table \ref{tab:entity_recognition_results}. ER evaluation was performed with three random seed initialization 3, 37 and 42 on the HuggingFace token classification script.

\begin{table}[h]
\begin{center}
    \begin{tabular}{l|l|l|l}
        \hline
        \textbf{model} & \textbf{batch} & \textbf{lr} & \textbf{epoch} \\
        \hline
       bert-base & 16 & 5e-5 & 10 \\
        bert-base+CRF & 32 & 1e-4 & 5 \\
        bert-large& 64 & 1e-4 & 10 \\
        bert-large+CRF & 16 & 5e-5 & 10 \\
        deberta-base & 32 & 5e-5 & 5 \\
        deberta-base+CRF & 32 & 5e-5 & 5 \\
        jobbert& 32 & 5e-5 & 10 \\
    jobbert+CRF & 32 & 1e-4 & 5 \\
        roberta & 32 & 1e-4 & 5 \\
        roberta+CRF & 16 & 1e-4 & 5 \\
        roberta-large & 32 & 5e-5 & 5 \\
        roberta-large+CRF & 32 & 5e-5 & 5 \\
        ESCOXLM-R & 32 & 5e-5 & 5 \\
        ESCOXLM-R+CRF & 32 & 5e-5 & 5 \\
        \hline
    \end{tabular}
    \captionof{table}{Entity Recognition best training hyperparameters}
    \label{tab:erhyper}
\end{center}
\end{table}

\subsubsection*{Entity Similarity}
For the entity similarity training, we used the sbert official website to train our models. We resulted on training with the default parameters without hyperparmeter search. In Table \ref{tab:training-config}, we present the hyperparameters.
\begin{table}
    \centering
    \begin{tabular}{ll}
        \toprule
        \textbf{Parameter} & \textbf{Value} \\
        \midrule
        Epochs & 2 \\
        Evaluation Steps & 0 \\
        Evaluator & NoneType \\
        Max Gradient Norm & 1 \\
        Optimizer Class & AdamW \\
        Learning Rate (lr) & 2e-05 \\
        Scheduler & WarmupLinear \\
        Warmup Steps & 10000 \\
        Weight Decay & 0.01 \\
        \bottomrule
    \end{tabular}
\caption{Summary of the training configuration parameters for Sentence Transformers}
    \label{tab:training-config}
\end{table}

\subsection*{Prompts used in this study}
\label{sec: prompts}
The following prompts have been used in the Datasets section to judge between the qualification annotations.

\textbf{Prompt: }"In the context of the following sentence choose the appropriate EQF level that suits the qualification.
If you cannot determine the EQF level answer UNK. \newline
Example: \newline
Sentence: Qualifications and experiences : BSc , MSc or PhD or equivalent in Statistics , Computer Science , Mathematics or other analytical field .\newline
Qualification: BSc , MSc or PhD or equivalent in Statistics , Computer Science , Mathematics or other analytical field .\newline
EQF level: EQF8\newline
Sentence: $\{$sentence$\}$\newline
Qualification: $\{$qualification$\}$\newline
EQF level: "

The following prompts have been used in the Transformer Decoder Integration section to generate synthetic queries from Occupations and Skills datasets for evaluation. For each datapoint, the prompt is adapted depending on the original job title, job description or skill description.

\textbf{Occupation prompt}: "Given the following description of the user's past job, return the answer of the user to the following question.

Description: $<$\textit{title}$>$ $<$\textit{description}$>$

Question: Describe your last job. Answer in one sentence. Don't be too formal.

Answer:"

\textbf{Skill prompt}: "Given the following description of the user's skill, return the answer of the user to the following question.
Description: $<$\textit{description}$>$
Question: What are your skills and expertise? Answer in one sentence. Don't be too formal.
Answer:"

Lastly we present the prompt template used in section \ref{sec:decoders} to perform entity extraction with transformer decoders.

\textbf{Prompt:} "A virtual assistant answers questions from a user based on the provided text.
\newline
\$few shots\$
\newline
USER: Text: \$text\$
\newline
ASSISTANT: I’ve read this text.
\newline
USER: What describes \$entity\$ in the text?
\newline
ASSISTANT: "
\newline
where we replace the few shot, text and entity placeholders with data points from the Green Benchmark.

\subsection*{Dataset tables}
\label{sec:tables}
All analysis in this section was done with the NLTK \footnote{https://www.nltk.org/} package.

\begin{table}[h]
    \centering
    \begin{tabular}{c|l|l}
        \toprule
        \multirow{5.3}{*}{\rotatebox[origin=c]{90}{\textbf{\scriptsize TRAIN}}} &  & \textbf{Statistics} \\
        \midrule
        & Sentences & 9,634 \\
        & Tokens & 233,628 \\
        & Entity Spans & 18,098 \\
        \midrule
        \multirow{3}{*}{\rotatebox[origin=c]{90}{\textbf{\scriptsize TEST}}}

        & Sentences & 336 \\
        & Tokens & 8,050 \\
        & Entity Spans & 904 \\
        \midrule
        & Average Entity Length & 3.67 \\
        \bottomrule
    \end{tabular}
        \captionof{table}{\textbf{Green Benchmark} Data Analysis}
        \label{tab:green_benchmark}
\end{table}

\begin{table}[h]
\centering
    \begin{tabular}{l|l}
        \toprule
         \textbf{EQF Level} & \textbf{Statistics} \\
        \midrule
         1 & 40 \\
         2 & 88 \\
         3 & 89 \\
         4 & 166 \\
         5 & 115 \\
         6 & 128 \\
         7 & 117 \\
         8 & 74 \\       
        \midrule
         Total & 814 \\
        \hline
         Average Word Length & 7.24 \\
         \midrule
         Total Countries & 30\\
         Entries per country & 27.13 \\
        \bottomrule
    \end{tabular}
        \captionof{table}{\textbf{EQF reference database} Data Analysis}
        \label{tab:eqf_data_analysis}
\end{table}



\end{document}